\documentclass{article}

\usepackage{amsmath, amssymb}  
\usepackage{graphicx}          
\usepackage{booktabs}

\usepackage{enumitem}
\usepackage{microtype}
\usepackage{wrapfig}
\usepackage{adjustbox}

    \usepackage[preprint]{neurips_2025}

\usepackage[utf8]{inputenc} 
\usepackage[T1]{fontenc}    
\usepackage{hyperref}       
\usepackage{url}            
\usepackage{booktabs}       
\usepackage{amsfonts}       
\usepackage{nicefrac}       
\usepackage{microtype}      
\usepackage{xcolor}         

\title{LoRACode: LoRA Adapters for Code Embeddings}

\author{%
  Saumya Chaturvedi \\
  Max Planck Institute for Software Systems\\
  Saarbrücken, Germany \\
  \texttt{schaturv@mpi-sws.org} \\
  \And
  Aman Chadha \\
  AWS GenAI \\
  Santa Clara, CA, USA \\
  \texttt{hi@aman.ai} \\
  \AND
  Laurent Bindschaedler \\
  Max Planck Institute for Software Systems \\
  Saarbrücken, Germany \\
  \texttt{bindsch@mpi-sws.org} \\
}

\begin{document}

\maketitle

\begin{abstract}
  Code embeddings are essential for semantic code search; however, current approaches often struggle to capture the precise syntactic and contextual nuances inherent in code. Open-source models such as CodeBERT and UniXcoder exhibit limitations in scalability and efficiency, while high-performing proprietary systems impose substantial computational costs. We introduce a parameter-efficient fine-tuning method based on Low-Rank Adaptation (LoRA) to construct task-specific adapters for code retrieval. Our approach reduces the number of trainable parameters to less than two percent of the base model, enabling rapid fine-tuning on extensive code corpora (2 million samples in 25 minutes on two H100 GPUs). Experiments demonstrate an increase of up to 9.1\% in Mean Reciprocal Rank (MRR) for Code2Code search, and up to 86.69\% for Text2Code search tasks across multiple programming languages. Distinction in task-wise and language-wise adaptation helps explore the sensitivity of code retrieval for syntactical and linguistic variations. To foster research in this area, we make our code and pre-trained models publicly available\footnote{\href{https://github.com/loracode-submission/loracode}{Github code repository}, \href{https://huggingface.co/collections/anonymous-loracode-submission/loracode-models-682644ebc51069da69e329eb}{Huggingface models trained for LoRACode}}.
\end{abstract}

\section{Introduction}

Code search based on natural language and code-based queries has long been recognized as a core problem in software engineering, and was in fact, one of the earliest tasks tackled by AI methods. \citep{codesearchrelevance, codedomainsurvey, dlcodesurvey} denotes the immense applications of probabilistic code search models in recommender systems, inferring code convention, code translation and other similar domains. Over time, the proliferation of open source repositories and question-and-answer forums (e.g., StackOverflow) has only underscored the importance of semantic code search. Modern systems therefore seek rich semantic representations of code; indeed, recent embedding-based retrieval frameworks (for example, Voyage AI’s \citep{voyage} code embedding model and OpenAI’s code embeddings \citep{openaiembeddings}) have achieved dramatic gains in search performance, yet stay closed-source and cannot be fine-tuned.

On the other hand, Transformer-based encoders like CodeBERT \citep{codebert}, GraphCodeBERT \citep{graphcodebert}, UniXcoder \citep{unixcoder}, and StarCoder \citep{starcoder}, while effective in learning deep joint representations of code and text, are often smaller in scale and based on architectures like BERT, limiting their performance in tasks such as the semantic retrieval of code snippets across text and code-based queries. These models also fail to generalize over cross-lingual tasks, and hence perform poorly on complex tasks than their variants fine-tuned for specific downstream applications \citep{dlcodesurvey}. With large web-scraped datasets containing about 2 million samples \citep{csn}, fine-tuning also becomes highly resource-intensive.




We present a novel approach to code search by introducing Parameter-Efficient Fine-Tuning (PEFT) methods, specifically Low-Rank Adaptation (LoRA)~\citep{lora}. This method aims to create \emph{task-specific and language-specific adapters} for retrieving code snippets. By significantly reducing the number of trainable parameters, it enables the fine-tuning of large-scale models \emph{with minimal computational resources} while achieving state-of-the-art (SOTA) performance. Our approach operates efficiently, requiring only 1.83\% to 1.85\% of the parameters used in the base models for fine-tuning. Furthermore, it can be trained on \textbf{2 million code samples} in just \textbf{25 minutes} using two H100 GPUs. This improvement leads to significant enhancements in the Mean Reciprocal Rank (MRR@1) for both Code2Code and Text2Code search tasks. Finally, we propose creating adapters that encapsulate language-specific features across six programming languages, fine-tuning them separately to achieve significant improvements in Mean Reciprocal Rank for Text2Code search.

We compare our proposed method to existing models used in code search tasks across various programming languages. We evaluate performance based on accuracy and retrieval efficiency, organizing the training of the adapters based on task-specific and language-specific capabilities. Our findings demonstrate that our approach outperforms current systems while also reducing computational costs.

Our key contributions can be summarized as follows:
\begin{itemize}
    \item Introduction of a novel parameter-efficient fine-tuning (PEFT) approach for code search utilizing Low-Rank Adaptation (LoRA).
    \item Efficient fine-tuning that employs only 1.83\%–1.85\% of the parameters used in base models, significantly improving computational efficiency.
    \item Proposing the use of language-specific adapters for Text-to-Code retrieval tasks and evaluating the corresponding performance improvement across six programming languages.
    \item An increase of up to 9.1\% in Mean Reciprocal Rank (MRR@1) for Code2Code and up to 86.69\% for Text2Code retrieval tasks.
\end{itemize}

The rest of the paper describes our approach. Section~\ref{sec:related-work} provides background information and related work that motivates our design. Section~\ref{sec:implementation} details our approach and its implementation. Section~\ref{sec:evaluation} evaluates the efficiency and accuracy of our solution and compares it with the state-of-the-art. Finally, Section~\ref{sec:conclusions} provides our conclusions. 

\begin{figure}[t]
    \centering
    \includegraphics[width=\linewidth]{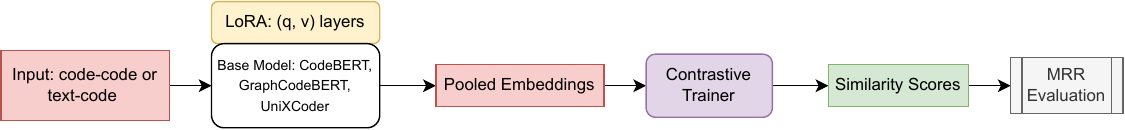}
    \caption{LoRACode Architecture: The input consists of code-code or text-code pairs. The base models are enhanced with LoRA layers, that output pooled embeddings, optimized using the Contrastive Trainer to improve retrieval accuracy. Finally, Mean Reciprocal Rank (MRR) measures the quality of ranked retrieval results.}
    \label{fig:loracode-architecture}
\end{figure}

\section{Background \& Related Work}
\label{sec:related-work}
\textls[-10]{This section provides an overview of related work in the field of code search, covering key embedding models, benchmarks, and techniques. We first discuss various code embedding models, highlighting their strengths and limitations, followed by an exploration of relevant datasets and benchmarks for evaluating code retrieval tasks. The section also delves into techniques such as LoRA and contrastive fine-tuning, which are crucial for enhancing model performance with minimal computational resources.}

\subsection{Code Search and Embedding Models}
Earlier research \citep{metacoderetrieval, bimodalmodeling} imagined natural language query-based code search as a statistical modeling problem free from the application of Transformers. Neural Code Search \citep{metacoderetrieval} extracted keywords from function, variable names and comments, and used skip-gram vectorization to create token and document embeddings whose cosine similarity is computed against query embeddings to facilitate search.

While previous work managed to infer merely surface-level features from code tokens and names, CodeBERT \citep{codebert} took a revolutionary dive into learning deep code semantics by introducing a bimodal pre-trained BERT model for programming and natural languages. Trained using Masked Language Modeling and Replaced Token Detection on both source code and natural language, CodeBERT \citep{codebert} leveraged transformer attention mechanisms to model relationships, inviting scope for many more code embedding models possessing the ability to be fine-tuned on a variety of downstream applications. GraphCodeBERT \citep{graphcodebert} extends CodeBERT by incorporating a data flow graph to capture structural dependencies in code. This enhancement enables the model to better understand variable usage and control flow, making it more effective for bug detection and code summarization tasks. UniXcoder \citep{unixcoder} unifies code representation across multiple modalities, including text, code, and structured representations. By leveraging cross-modal pretraining, it achieves state-of-the-art results in tasks like code translation and completion. 

StarCoder \citep{starcoder} is a transformer-based model trained on over 80 programming languages, known for its high-quality code completions. However, its heavy computational demands and proprietary training data limit open-source use. CodeT5 \citep{codet5} features a unified framework for code understanding and generation, utilizing a pre-trained encoder-decoder Transformer model to enhance multi-task learning. CodeT5+ \citep{codet5+} builds on frozen large language models, avoiding the need for training from scratch, and has been tested across more than 20 code-related benchmarks in various settings.

Despite the advancements in code encoding made by Transformer-based models \citep{dlcodesurvey}, they fail to generalize to different tasks and show sub-par performance in basic retrieval tasks due to their lightweight architecture. The semantic gap between natural language and code contributes to the complexity added by the diversity of code representation by programming languages. \citet{lace} show that previous code embedding models struggle to separate language-specific semantic components from language-agnostic syntactic components during search evaluations. \citep{lace} employs methods to eliminate language-specific information, leading to significant performance improvements in retrieval tasks by focusing on language-agnostic components.

\subsection{Benchmarks and Datasets}
CodeXGLUE \citep{codexglue} is a benchmark for code understanding and generation, encompassing tasks such as code-to-text generation, code completion, and code translation. It offers pre-processed datasets and evaluation metrics, making it essential for benchmarking code models.

XLCost \citep{zhu2022xlcost} is a benchmark dataset created from GeeksForGeeks, utilizing the same function to create parallel code snippets across seven programming languages. CosQA \citep{cosqa} is a question-answering dataset with human-annotated query-code pairs, ensuring strong semantic alignment for text-to-code search tasks. While effective for fine-tuning models, its focus on Python limits its applicability to other programming languages.

The Code Information Retrieval (CoIR) benchmark \citep{coir} evaluates small and large-scale retrieval tasks to assess model performance across different computational needs. It reveals a performance gap between open-source models and proprietary solutions like OpenAI's embeddings, emphasizing the need for efficient retrieval systems.

\subsection{Low-Rank Decomposition}
LoRA \citep{lora} has emerged as a parameter-efficient fine-tuning technique for large language models. It reduces the need for full model fine-tuning by introducing low-rank decomposition matrices into the attention layers, leaving the pre-trained weights frozen. This approach significantly lowers the memory and computational requirements, making adapting large models for downstream tasks feasible even with constrained resources. LoRA's modular design allows task-specific adaptation without compromising the integrity of the base model, significantly reducing the trainable parameter count to as low as 1\%-2\% of the total model parameters.

Jina AI's \texttt{jina-embeddings-v3} \citep{jina} employs LoRA adapters for  multilingual and long-context retrieval. By fine-tuning low-rank matrices, it generates high-quality embeddings for query-document retrieval and text matching with minimal computational cost, demonstrating LoRA's efficiency for enhancing code embeddings in text-to-code and code-to-code retrieval.

\subsection{Embeddings and Contrastive Fine-tuning}
Improving text embeddings is a key focus, especially for smaller language models. Techniques like contrastive fine-tuning \citep{contrastive} enhance embeddings by aligning semantically similar text pairs, using contrastive and triplet loss \citep{contrastive-loss,triplet-loss} from Computer Vision. LoRA further refines these embeddings efficiently by introducing low-rank matrices into the attention layers, allowing customized adjustments for specific datasets or tasks \citet{lora} .

Studies \citep{contrastive-text} showed that contrastive learning frameworks enhance sentence and document embeddings by minimizing distances between positive pairs and maximizing distances between negative ones. Integrating LoRA with these frameworks provides a lightweight yet effective way to fine-tune models for semantic similarity, text retrieval, and classification tasks.

\citet{contrabert} discusses a contrastive pre-training task involving nine data augmentation operators that transform original program and natural language sequences. The variants, when paired with the original sample, enhance token representations and model robustness.

\citet{contrastive-text-code} trains text and code embedding models separately using a contrastive learning objective with in-batch negatives on unlabelled data. The text models were fine-tuned on neighboring word pairs from the internet, while the code models focused on \texttt{(text, code)} pairs from CodeSearchNet \citep{csn}, achieving a 20.8\% improvement over previous work through unsupervised contrastive fine-tuning.

The authors of \citep{peft-embeddings} mention the usage of PEFT methods like LoRA and Prompt Tuning for Text2Code search but do not document any information on the implementation or the analysis of their low-rank adaptation approaches.

\citet{codexembed} highlight advancements in code retrieval using instruction-tuned large language models (LLMs) and evaluates performance with the NDCG@10 metric across eight tasks. It points out the lack of large, open-source embedding models for code retrieval, as most are proprietary. CodeXEmbed outperforms Voyage Code \citep{voyage} on the CoIR benchmark \citep{coir} by 20\% and introduces various retrieval models. However, its large model reliance incurs high storage and computation costs. Our work addresses this by utilizing LoRA adapters on smaller code embedding models, significantly reducing resource overhead while improving performance.

\section{Design and Implementation}
\label{sec:implementation}

LoRACode leverages LoRA adapters for code search tasks. We create fine-tuned adapters on two primary functions for source-included multilingual search: Text-to-Code (Text2Code) search and Code-to-Code (Code2Code) search. For these tasks, the model processes code tokens and docstring tokens to convert them into embeddings, computes a similarity score for each query point, and calculates the Mean Reciprocal Rank over the sorted array of similarity scores of the relevant code programs. The backbone of our system is a pre-trained code embedding model, such as CodeBERT \citep{codebert}, GraphCodeBERT \citep{graphcodebert}, or UniXcoder \citep{unixcoder}, which is fine-tuned with LoRA \citep{lora}. LoRA \citep{lora} enables efficient fine-tuning of these models (or a combination of them) by introducing low-rank adaptation matrices into the attention layers while freezing the rest of the model’s parameters. These adapters are categorized based on their capabilities into Text2Code and Code2Code adapters and are further divided by language capabilities across six programming languages for Text2Code search.

We utilize \textit{ContrastiveTrainer} \citep{contrastive}, a custom extension of the Hugging Face Trainer class designed for efficient contrastive learning. This trainer minimizes a cosine similarity-based loss function between query and positive code embeddings, improving retrieval accuracy. Furthermore, we implemented pooled embeddings by averaging hidden states across the sequence length while excluding padding tokens. This was achieved by modifying the attention mask to exclude padded tokens from computation. This pooling strategy ensures that the final embeddings retain meaningful contextual representations by aggregating token-level features while mitigating the influence of padding tokens, crucial for maintaining semantic integrity across variable-length code snippets. 

Figure \ref{fig:loracode-architecture} showcases the simple sequence flow followed by LoRACode from accepting query-code pairs all the way to Mean Reciprocal Rank (MRR) Evaluation. Tables~\ref{tab:training_hyperparams} and \ref{tab:lora_hyperparams} summarize the training and LoRA hyperparameters used in our experiments respectively.

\textls[-8]{Each pretrained model -- CodeBERT \citep{codebert}, GraphCodeBERT \citep{graphcodebert}, and UniXcoder \citep{unixcoder} -- was fine-tuned using the LoRA configuration. For a detailed explanation of the choice of models, see Appendix \ref{models-used}. LoRA employs low-rank decomposition to modify only the attention layers while freezing the remaining weights. This approach reduces memory consumption and accelerates training.
}
\begin{table}[h]
    \centering
    \begin{minipage}{0.48\linewidth}
        \centering
        \caption{Training hyperparameters}
        \vspace{1mm}
        \label{tab:training_hyperparams}
        \footnotesize
        \begin{tabular}{ll}
            \toprule
            \textbf{Hyperparameter} & \textbf{Value} \\
            \midrule
            Batch Size & 16 per device \\
            Epochs & 1 (rapid evaluation) \\
            Learning Rate Scheduler & 1000 warmup steps \\
            Logging & Every 200 steps \\
            Save Strategy & End of each epoch \\
            Evaluation Strategy & No intermediate eval \\
            \bottomrule
        \end{tabular}
    \end{minipage}
    \hspace{1mm}
    \begin{minipage}{0.42\linewidth}
        \centering
        \caption{LoRA hyperparameters}
        \vspace{1mm}        
        \label{tab:lora_hyperparams}
        \footnotesize
        \begin{tabular}{ll}
            \toprule
            \textbf{Hyperparameter} & \textbf{Value} \\
            \midrule
            Ranks & 16, 32, 64 \\
            LoRA Alpha & 32, 64, 128 \\
            Target Modules & Query and Value \\
            Dropout & 10\% \\
            \bottomrule
        \end{tabular}
    \end{minipage}
\vspace{-1mm}
\end{table}

In various experiments conducted with StarCoder \citep{starcoder}, we tested different combinations of the Query, Key, and Value layers within the attention mechanism. This was done to analyze the contributions of the early, middle, and outer layers in low-rank decomposition. For more details about the training procedure, please refer to Appendix \ref{training-procedure}.

\subsection{Task Specific Adapters}
The task-based approach for fine-tuning utilizes a custom \textit{ContrastiveTrainer} to train the model for either Text2Code or Code2Code retrieval tasks. For Code2Code search, the data loader constructs pooled embeddings for both query and relevant code snippets, incorporating language-specific features. Conversely, for Text2Code search, features are generated for the docstring along with anchor code snippets. For each query sample, similarity scores are computed for all retrieval candidates, and the resulting embedding vectors are then sorted. These sorted vectors are assigned ranks, and the Mean Reciprocal Rank (MRR) is calculated as the average of $\frac{1}{\text{Rank}}$ across all queries.

\subsection{Language Specific Adapters}
The language-based approach focuses on tailoring the model to effectively manage the unique syntax and semantics of various programming languages. Instead of using a single adapter for all languages, we develop language-specific adapters by fine-tuning the model on datasets tailored to each programming language. This process involved separating the language datasets using CodeSearchNet \citep{csn} to create data loaders, as well as training and testing datasets for samples pertaining to a specific programming language. The feature construction, fine-tuning, and mean reciprocal rank (MRR) evaluation remain unchanged.

\section{Evaluation}
\label{sec:evaluation}

In this section, we conduct a comprehensive evaluation of LoRACode using a multifaceted approach. This includes examining different datasets, various methods of training the LoRA matrices for distinct tasks, and a range of programming languages. Our primary focus is on analyzing accuracy metrics to assess the efficacy of LoRA adapters in code retrieval tasks. We aim to provide detailed answers to the following research questions:

\begin{enumerate}[label=\textbf{RQ\arabic*:}]
    \item How does the performance of LoRACode, measured through Mean Reciprocal Rank and Normalized Distributed Cumulative Gain, compare to large pre-trained code embedding models (CodeBERT, UniXcoder, GraphCodeBERT) in code retrieval tasks? (\S\ref{subsec:eval-code2code}, \S\ref{subsec:eval-text2code})
    \item To what extent does using LoRA's low-rank decomposition in the attention layers reduce the computational cost and memory consumption while maintaining or improving retrieval performance on multilingual code search tasks?  (\S\ref{subsec:eval-code2code})
    \item What is the impact of fine-tuning code retrieval models using language-specific adapters versus task-specific adapters across different programming languages? (\S\ref{subsec:language-specific})
\end{enumerate}

\subsection{Experimental Setup}
\label{subsec:experimental-setup}

\subsubsection{Datasets}
For Text2Code retrieval, we utilized the CodeSearchNet dataset \citep{csn}, containing over 2 million methods from open-source GitHub projects in six languages: Go, Java, JavaScript, PHP, Python, and Ruby. Each method includes natural language documentation (e.g., docstrings) and metadata such as repository, location, and line numbers. We also fine-tuned for Text2Code using the CosQA \citep{cosqa} dataset, a question-answering dataset for Python code tokens. For Code2Code retrieval, we employed the XLCost dataset \citep{zhu2022xlcost}, parallel across seven languages (C++, Java, Python, C\#, JavaScript, PHP, and C) at both the snippet and program levels. Programs are divided into snippets, maintaining alignment across languages.

We merged the datasets across all languages for each independent task and removed duplicate queries with identical source identifiers (across original programming languages). The data was tokenized and corrected using language-specific adaptations to handle unique tokens (e.g., \texttt{NEW\_LINE}). Data collators were then employed to create batches of query-relevant code sequences for the Code2Code search and anchor-docstring pairs for the Text2Code search with appropriate labels.

\subsubsection{Metrics}
The evaluation metric used for these experiments was Mean Reciprocal Rank (MRR). MRR@$K$ measures the system's effectiveness in identifying relevant results as the top-ranked output. This metric focuses on the order of the first \( K \) relevant results, disregarding the number or order of subsequent results. We also chose the Normalized Discounted Cumulative Gain@10 metric to evaluate the efficiency of the text-based retrieval since some papers \citep{coir} cite the importance of NDCG in not only considering the order of retrieved items but also their relevance intensity.

\subsubsection{Hardware \& Configuration} 
We perform all experiments on a machine with two H100 GPUs equipped with 80 GB of HBM each, configured with Debian Ubuntu, running PyTorch version 2.5.1 and Transformers version 3.5.0.

\subsection{Text2Code}
\label{subsec:eval-text2code}

For Text2Code search, we fine-tune the base code embedding models UniXcoder, GraphCodeBERT, and CodeBERT over the CodeSearchNet \citep{csn} dataset with the same varying LoRA ranks and similar creation of combined adapter.

Table~\ref{text2code-csn} shows the Mean Reciprocal Ranks received on training LoRA adapters of ranks 16, 32, 64 trained on the combination of CodeBERT \citep{codebert}, GraphCodeBERT \citep{graphcodebert} and UniXCoder \citep{unixcoder}.

\begin{table}
  \centering
  \begin{tabular}{ccccccccc}
   \toprule
    & \textbf{Ruby} & \textbf{Go} & \textbf{PHP} & \textbf{Python} & \textbf{Java} & \textbf{Javascript} \\
    \toprule
  LoRACode - Combined (rank 64) & 42.83 & 48.34 & 20.88 & 28.60 & 33.08 & 30.55 \\
  LoRACode - Combined (rank 32) & 43.96 & 48.98 & 21.86 & 29.85 & 34.15 & 31.38 \\
  UnixCoder & 44.06 & 49.59 & 22.31 & 29.76 & 34.47 & 32.05 \\
  GraphCodeBERT & 20.80 & 12.48 & 8.08 & 10.38 & 8.60 & 7.30 \\
  CodeBERT & 0.37 & 0.15 & 0.03 & 0.06 & 0.04 & 0.06 \\
  Starencoder & 4.41 & 1.85 & 0.57 & 2.14 & 1.89 & 1.55 \\
  \bottomrule
  \end{tabular}
  \caption{\label{text2code-csn}
    MRR results for Text2Code search of LoRA models UniXCoder, GraphCodeBERT, and CodeBERT fine-tuned with ranks 32 and 64, compared with base models, evaluated over XLCost dataset per language. LoRACode Combined denotes a single adapter of given rank fine-tuned over 3 base models: UniXcoder, GraphCodeBERT, and CodeBERT, but evaluated over UniXcoder. \vspace{-5mm} }
\end{table}

The results do not show a substantial increase in the MRR for the LoRA models compared to the UniXCoder base model. At best, LoRA configurations of rank 32 and lower, while performing better than GraphCodeBERT, CodeBERT, and StarCoder, perform at par with UniXCoder. Higher ranks report lower MRR scores. This indicates that a task-wise breakdown of the fine-tuning requirement does not translate equally for text-based retrieval as it did for code-based retrieval. 

To explore the second question highlighted in Section \ref{sec:evaluation}, we fine-tune the UniXcoder model for Text2Code retrieval on a single programming language. We use the CosQA \citep{cosqa} dataset, as highlighted in Section \ref{subsec:experimental-setup}. Since CosQA is a question-answering dataset, we experimented with setting the task type for the LoRA Configuration \citep{houlsby2019parameterefficienttransferlearningnlp} \citep{lora} as \texttt{QUESTION\_ANS}, which was not feasible for retrieval tasks, since question-answering is a generative task, and would also require the inference using special Roberta \citep{roberta} models encased with question answering, and those suitable for handling parameters like start\_position, end\_position, etc. The task type was then set to \texttt{FEATURE\_EXTRACTION}, the model parameters were frozen, and then the LoRA matrices were trained using Contrastive Trainer. Detailed observations for the same are noted in Appendix \ref{training-procedure}.

\begin{wrapfigure}{r}{0.55\linewidth}
  \centering
  \begin{adjustbox}{width=\linewidth}
    \begin{tabular}{lcc}
      \toprule
       & \textbf{UniXCoder} & \textbf{LoRACode - Combined ($r$=64)} \\
      \toprule
      MRR & 31.36 & \textbf{36.02} \\
      NDCG@10 & 35.64 & \textbf{40.44} \\
      \bottomrule
    \end{tabular}
  \end{adjustbox}
  \caption{Increase in Mean Reciprocal Rank and Normalized Discounted Cumulative Gain @ k=10 for LoRACode combined adapter (rank 64) trained on CosQA dataset \citep{cosqa}, compared to the UniXCoder base model. LoRACode Combined denotes a single adapter fine-tuned over the 3 base models (UniXcoder, GraphCodeBERT, and CodeBERT), but evaluated on UniXcoder. \vspace{-5mm}}
  \label{text2code-cosqa}
\end{wrapfigure}

Table~\ref{text2code-cosqa} showcases the Mean Reciprocal Rank and Normalized Discounted Cumulative Gain@10 recorded for the combined adapter of rank 64 trained on the CosQA dataset \citep{cosqa} when compared to the UniXCoder base model.

There is an increase of 14.8\% in MRR and 13.5\% in NDCG for Text2Code fine-tuned over CosQA \citep{cosqa} dataset, as opposed to the CodeSearchNet \citep{csn} dataset. The observed performance gain can be attributed to the following key factors:

\begin{itemize}
    \item \textbf{Dataset Size}: CosQA \citep{cosqa} is a smaller dataset (20k samples) than CSN (2 million samples) \citep{csn}. While training on a smaller dataset generally leads to better accuracy, this alone is not the primary driver of the observed improvement.

    \item \textbf{Human-Annotated Data}: CosQA \citep{cosqa} is a human-annotated dataset designed specifically for question-answering tasks. The queries in CosQA \citep{cosqa} are highly coherent, as the corresponding code snippets directly address the text queries with minimal noise. Each sample in the dataset is also labeled (0/1) to indicate whether the given snippet correctly answers the query.
    
    \item \textbf{Programming Language-Specific Context}: CosQA \citep{cosqa} contains only Python code snippets, whereas CodeSearchNet \citep{csn} spans seven programming languages. In our experiments, task-specific adapters were employed instead of language-specific adapters. This likely diluted the results, as training a single adapter on a single language facilitates learning more nuanced features, syntax, and contextual information. In contrast, multilingual datasets like CodeSearchNet \citep{csn} introduce diverse language contexts, which may reduce the efficacy of adapters due to their limited trainable parameters. This hypothesis aligns with findings in previous papers \citep{lace}, which report lower results for multilingual search than monolingual search, albeit without analyzing language-specific features.
\end{itemize}


For these reasons, we found it appropriate to create programming language-specific adapters. MRR and NDCG showed massive improvements during the evaluation of language-specific adapters, as opposed to adapters fine-tuned on the combination of the dataset and the removal of duplicates. The results are encapsulated in Table \ref{text2code-language-specific}. 

\begin{table}
  \setlength{\tabcolsep}{6pt}
  \centering
  \begin{tabular}{cccccccc}
   \toprule
    \textbf{Languages} & \multicolumn{2}{c}{\textbf{MRR}} & \multicolumn{2}{c}{\textbf{NDCG}} & \textbf{\# Samples} & \textbf{Train Time} & \textbf{\% Increase} \\
     & \textbf{Base} & \textbf{LoRA} & \textbf{Base} & \textbf{LoRA} & & & \\
    \toprule
    Ruby & 44.06 & \textbf{45.78} & 47.95 & \textbf{49.77} & 1558 & 7:01 & 3.90 \\
    Go & 49.59 & \textbf{82.88} & 53.66 & \textbf{85.35} & 10456 & 47:36 & 67.13 \\
    PHP & 35.22 & \textbf{52.46} & 24.73 & \textbf{56.54} & 15078 & 1:08:39 & 48.94 \\
    Python & 29.76 & \textbf{55.56} & 32.83 & \textbf{59.49} & 15739 & 2:10:39 & 86.69 \\
    Java & 34.47 & \textbf{53.47} & 37.94 & \textbf{57.45} & 10308 & 1:25:19 & 31.91 \\
    Javascript & 32.05 & \textbf{38.75} & 35.04 & \textbf{42.35} & 3627 & 16:41 & 20.9 \\
    \bottomrule
   \end{tabular}
   \caption{Language-Specific Adapter Performance: MRR and NDCG Improvements Across Programming Languages. Across 6 programming languages, the Mean Reciprocal Rank and Normalized Discounted Cumulative Gain @ k=10 show significant improvements for the UniXCoder model adapted with language-wise adapters, compared to the base model. For comparison, the table details the number of code samples in the training dataset for each from CodeSearchNet, as well as the training time over 2xH100 GPUs and the \% increase in MRR. Training time is mentioned in minutes. \vspace{-5mm} 
}
   \label{text2code-language-specific}
\end{table}

These results demonstrate that training LoRA adapters on smaller, high-quality, and monolingual datasets like CosQA lead to substantial improvements in text-based retrieval performance. 

\subsection{Language Specific Experimentation}
\label{subsec:language-specific}


One noteworthy finding is that language-specific adapters perform better than task-specific adapters for the Text2Code retrieval task. Adapters trained on combined datasets that include multiple programming languages demonstrated reduced performance compared to those trained specifically for a single language. This performance difference can be attributed to:

\begin{itemize}
    \item Linguistic Diversity: Multilingual datasets introduce a wide variety of syntax, semantics, and contextual dependencies, diluting the model's ability to specialize in any one language.
    \item Limited Parameters: LoRA adapters have a restricted number of trainable parameters, making it challenging to generalize across diverse programming languages without sacrificing performance.
    \item Syntax and Structural Variations: Programming languages differ significantly in structure, such as Python's reliance on indentation versus Java's explicit use of braces. Task-specific adapters struggled to account for these differences.
\end{itemize}

The results highlight the importance of tailoring fine-tuning processes to individual programming languages rather than adopting a generalized approach. Some of the key insights shed light on the language-specific components of code search and fine-tuning using LoRA adapters, specifically the impact of training dataset size on the results:

\begin{itemize}
    \item When datasets were merged across languages, many duplicate samples (cross-lingual) were removed, drastically reducing the overall dataset size. This allowed training to be completed in just 16-20 minutes.
    \item In contrast, language-specific adapters were trained on datasets ranging from 10,000 to 20,000 samples per language (except JavaScript and Ruby, which were smaller). Training for a single language took over an hour, resulting in a richer learning process.
    \item Languages with larger datasets, such as Python (15,739 samples) and Go (10,456 samples), showed the highest performance improvements with increases in MRR of 86.69\% and 67.13\%, respectively.
    \item In contrast, Ruby (1,558 samples) and JavaScript (3,627 samples) exhibited much smaller improvements of 3\% and 20.9\%, respectively. This trend suggests a strong correlation between the dataset's size and the model's ability to learn effectively.
\end{itemize}

The baseline scores for language-specific adapters were consistent with prior works \citep{lace} \citep{coir} \citep{codexembed}. The scores were slightly lower in some cases, but the improvements with LoRA were significantly higher, validating the approach. The inherent differences in programming languages also contributed to the improvements. For instance:

\begin{itemize}
    \item Python's reliance on indentation and dynamic typing provided unique challenges that the language-specific adapter was better equipped to handle.
    \item Go, with its strict syntax and minimal redundancy, benefited from targeted fine-tuning that captured its simplicity and statically typed nature.
\end{itemize}

\subsection{Code2Code}
\label{subsec:eval-code2code}

In this section, we present the evaluation results of LoRA adapters for code retrieval tasks, focusing on the Mean Reciprocal Rank (MRR) performance across multiple programming languages. The table showcases the results of fine-tuning LoRA adapters at ranks 16, 32, and 64 on base models such as CodeBERT, GraphCodeBERT, and UniXCoder. The combined adapters are obtained by fine-tuning the same LoRA Config over the three models by loading it on a single model, freezing the model parameters, and saving the config to the HuggingFace hub. The adapter is evaluated by loading on top of UniXcoder \citep{unixcoder}. 

Table~\ref{code2code-mrr} shows the Mean Reciprocal Ranks received on training LoRA adapters of ranks 16, 32, 64 trained on the combination of CodeBERT \citep{codebert}, GraphCodeBERT \citep{graphcodebert} and UniXCoder \citep{unixcoder}. 

\begin{table}
  \centering
  \resizebox{\textwidth}{!}{%
  \begin{tabular}{ccccccccc}
    \toprule
    & \textbf{C} & \textbf{PHP} & \textbf{Java} & \textbf{C++} & \textbf{C\#} & \textbf{Javascript} & \textbf{Python}\\
  \toprule
  \textbf{LoRACode - Combined (rank 64)} & \textbf{41.07} & \textbf{44.18} & \textbf{48.84} & \textbf{48.91} & \textbf{48.69} & \textbf{48.56} & \textbf{48.27} \\
  LoRACode - Combined (rank 32) & 40.72 & 43.53 & 47.75 & 47.95 & 47.81 & 47.70 & 47.50 \\
  UnixCoder & 37.64 & 42.56 & 45.84 & 45.51 & 46.01 & 46.50 & 46.68 \\
  GraphCodeBERT & 32.81 & 37.93 & 30.86 & 34.04 & 31.74 & 39.53 & 20.30 \\
  CodeBERT & 27.45 & 30.47 & 24.80 & 10.54 & 25.53 & 25.56 & 5.48 \\
  Starencoder & 17.48 & 39.78 & 35.59 & 39.50 & 35.31 & 40.41 & 25.93 \\
  \bottomrule
  \end{tabular}
  }
  \caption{\label{code2code-mrr}
  MRR results for Code2Code search of the base models fine-tuned with LoRA modules at ranks 32 and 64, compared with base models. LoRACode Combined denotes a single adapter of given rank fine-tuned over UniXcoder, GraphCodeBERT and CodeBERT), but evaluated on UniXcoder. The average MRR across languages for LoRACode (rank 64) is 46.93 ± 2.28 (95\% confidence interval), details of the inference mentioned in Appendix \ref{statistical-significance}.\vspace{-5mm} 
  }
\end{table}


The LoRA adapters consistently outperformed embedding models like GraphCodeBERT and CodeBERT, demonstrating their efficacy in leveraging low-rank adaptations for code retrieval. Different languages report a significant increase in MRR, ranging from 9.1\% for C, 7.47\% for C++, 6.54\% for Java, 5.82\% for C\#, 4.43\% for Javascript, 3.40\% for Python, and 3.8\% for PHP. Using a LoRA config with rank 32 also substantially increases MRR over the highest-performing code embedding model UniXcoder \citep{unixcoder}. These findings suggest LoRA’s low-rank decomposition plays a crucial role in improving retrieval accuracy, supporting the use of these adapters for multilingual code search. The LoRA config utilizes only 1.83\% to 1.85\% of the total trainable parameters, leading to memory-efficient and inexpensive fine-tuning. 

A language-wise adaptation for this search task did not equal meaningful results, simply because the XLCoST \citep{zhu2022xlcost} dataset is designed to perform a cross-lingual code search through parallel samples, using the \texttt{function\_name} as the key identifier. See Appendix \ref{tables} for detailed observations regarding the time taken to generate embeddings for Code2Code search.

\section{Limitations}
\label{sec:limitations}
A key limitation of our approach lies in the narrow domain diversity of available training datasets. CodeSearchNet \citep{csn} and CosQA \citep{cosqa} are primarily built from GitHub code-comment pairs, reflecting open-source conventions and supporting only Text-to-Code retrieval. To enable code-to-code search, we rely on XLCost \citep{zhu2022xlcost}, which provides cross-lingual code pairs. While our experiments demonstrate the effectiveness of LoRA-based fine-tuning, we do not perform a deeper investigation of the exact language-specific representation in the Transformer architecture for these models. Additionally, large-scale LLM-based retrieval was left for future work due to computational constraints.

\section{Conclusion}
\label{sec:conclusions}
We introduced LoRACode, a parameter-efficient fine-tuning method based on Low-Rank Adaptation that significantly enhances code embeddings for both Text2Code and Code2Code retrieval tasks. Our experiments demonstrate substantial improvements in Mean Reciprocal Rank and Normalized Discounted Cumulative Gain across multiple programming languages while still maintaining low computational overhead. Our results indicate that language-specific adapters are superior to task-specific adapters in capturing the syntactic and semantic nuances of code. We plan to further investigate the parallels in language-specific adaptation for Code2Code search and across different languages.

\begin{ack}
    We would like to thank Saiteja Upatala for engaging conversations and helpful comments.
\end{ack}

\bibliographystyle{plainnat}
\bibliography{neurips_2025}

\clearpage
\newpage
\appendix
\renewcommand{\thesubsection}{A\arabic{subsection}}

\subsection{Models Used} \label{models-used}
We used the code embedding models CodeBERT, GraphCodeBERT, UniXcoder, and StarCoder as base models for parameter efficient fine-tuning. CodeBERT, GraphCodeBERT and UniXcoder were standardly available open-sourced models based on the BERT encoder model, whereas UniXcoder is a pre-trained unified encoder-decoder cross-modal model. RoBERTA based models are also straightforward tokenizers which make for easier fine-tuning. Starcoder is an open-sourced code LLM chosen for its extensive training on over 1 trillion tokens across 80+ programming languages.

CodeBERT demonstrated poor MRR scores because it is based on simple encoder model, whereas the encoder-decoder framework is sub-optimal for autoregressive tasks \citep{unixcoder}. This is why UniXcoder performed well as the base model, and also when adapted with LoRA addends.

\subsection{Training Procedure} \label{training-procedure}

    For fine-tuning a task-specific adapter, we loaded a \texttt{PEFTConfig} of desired rank, set \texttt{lora\_alpha} value to be double the rank, set dropout value as 0.1, set the target modules as the query value addends of the Attention layer and occasionally set the \texttt{task\_type} for the \texttt{PEFTConfig} to be \texttt{FEATURE\_EXTRACTION}. Some details of the implementation procedure and observations are noted as follows:

\begin{itemize}
    \item We noticed that the models performed better when set with the \texttt{FEATURE\_EXTRACTION} task flag for Text2Code search, but there weren't much significant improvements for Code2Code search task. This indicates the utility of task type in providing the hidden states that can be used as embeddings for appropriate feature or embedding for the downstream task in question. 
    \item We tried using a \texttt{task\_type} for \texttt{QUES\_ANSWERING} for the CosQA dataset \citep{cosqa} since it is specifically formatted for question answering tasks, but this would have required a special RobertaTokenizerForQuestionAnswering class during inference rather than the base abstraction RobertaTokenizer needed for MRR evaluation. So the \texttt{task\_type} was left at \texttt{FEATURE\_EXTRACTION}.
    \item Since Starcoder did not have specific query and value modules in its architecture, we discovered the layer-wise modules for each across the 12 layers. We experimented with different combinations of QV targets across the earlier, the middle and the later layers. We found that the aggregate of middle layers query value addends when used as target modules for LoRAConfig, performed better than the other two. The performance was still subpar for StarCoder due to LLMs being inefficient for code retrieval tasks and lack the ability to hold retrieval context during fine-tuning.
\end{itemize}

\subsection{Prevalent Knowledge} \label{prevalent-knowledge}
We learnt the calculation for last token pooling from similar papers. We thus calculated embeddings by not taking the average of last hidden states, but instead by reducing the masked layers, and only taking an average over the attention heads.

\begin{table}
  \centering
  \begin{tabular}{lcc}
    \toprule
     & \textbf{Base Model} & \textbf{PEFT Model} \\
    \toprule
     C &  0:52 & 0:44       \\
    PHP     &    4:30 & 3:44       \\
    Java     &    8:46 & 9:22       \\
    C++     &    8:41 & 9:27      \\
    C\#      &    8:37 & 9:24       \\
    Javascript     &  8:35 & 8:56       \\
    Python    &    8:24 & 9:17      \\
    \bottomrule
  \end{tabular}
  \caption{Time taken to generate embeddings (in MM:SS) for Code2Code search of the base embedding models vs embedding models altered with LoRA adapters. The latency is measured over each programming language's dataset, for the combined adapter of LoRA rank 64, over UniXCoder as base model. \vspace{-5mm} 
}
  \label{latency}
\end{table}

\subsection{Diagrams and tables} \label{tables}
Table \ref{latency} showcases the time taken to generate embeddings in minutes for Code2Code search of the base embedding models when compared with LoRA models. The latency is measured over each programming language's dataset, for the combined adapter of LoRA rank 64, over UniXCoder as base model. 

\begin{figure}[h]
\begin{center}
    \includegraphics[width=0.95\linewidth]{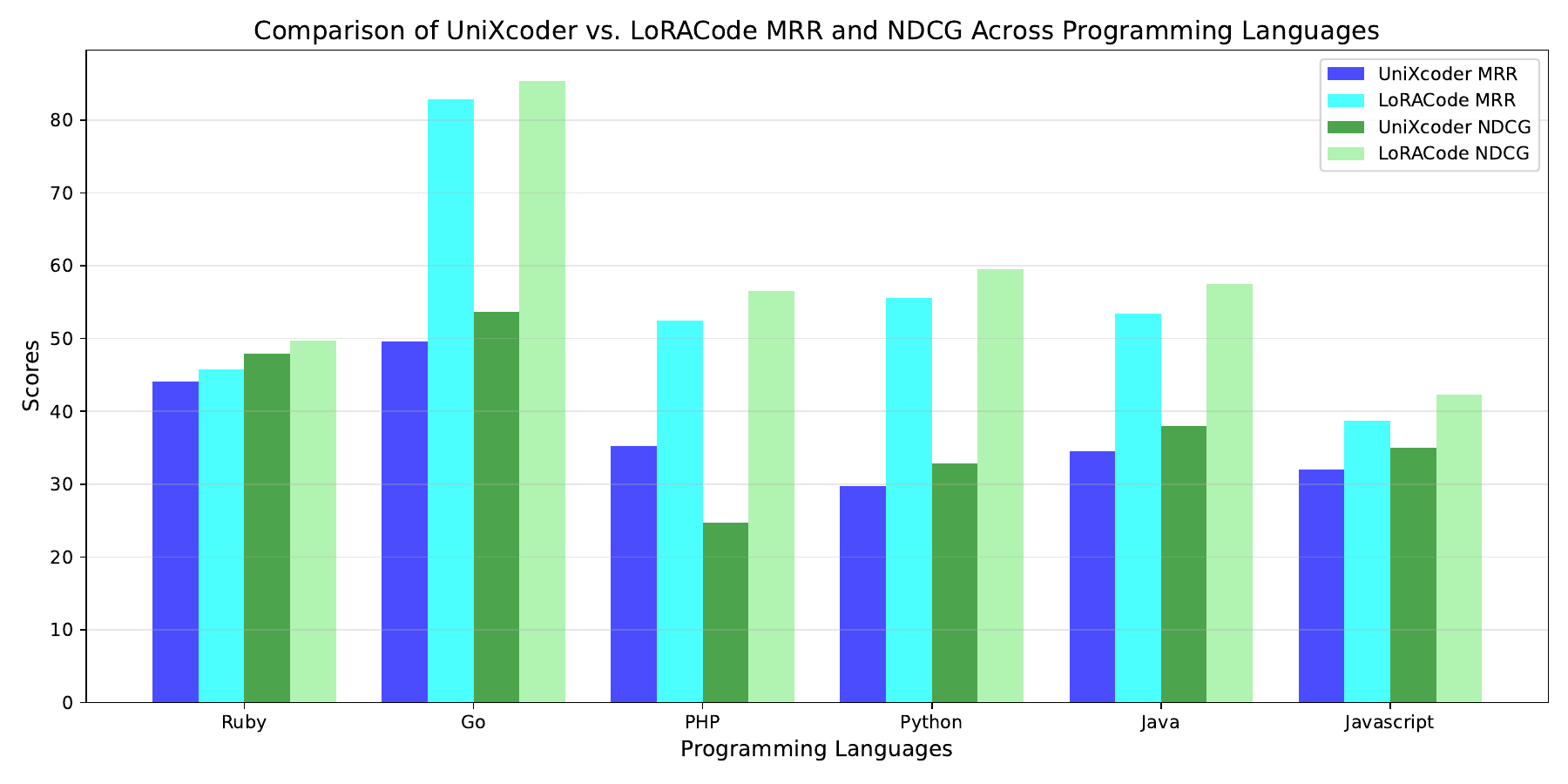}
\caption{Bar Graph showing the trends of MRR and NDCG for Base and LoRA models across different programming languages for Text2Code search.}
\label{bar-graph}
\end{center}
\end{figure}

Figure \ref{bar-graph} illustrates the trends in MRR and NDCG for Base models and LoRA-enhanced models across different programming languages. It shows that LoRA consistently improves both retrieval effectiveness (MRR) and ranking relevance (NDCG), with larger performance gains observed in languages with more training data. The visualization highlights the advantage of language-specific fine-tuning with LoRA, demonstrating significant improvements over the baseline models.

This bar graph shows a substantial increase in both MRR and NDCG, for LoRACode as opposed to the base UniXcoder model. Here LoRACode divided on language-specific adapters perform much better than LoRACode adapters fine-tuned on the aggregation of datasets across all languages, showcasing the importance of tailoring to linguistic diversity.

\subsection{Statistical Significance} \label{statistical-significance}
Code2code experiments report a mean MRR value of 46.93, with a 95\% confidence interval of ±2.28. Since code-based search does not emphasize cross-lingual importance, and a common base model (UniXCoder) was loaded with the same LoRA module (no language-wise adaptation) for fine-tuning and inference on all language datasets, we make this generalization for calculating statistical significance. In XLCoST, the datasets have an equal size for each language subset, so there is no variation in dataset size as well. For the case of Text2Code adapters, a similar generalization could not be conducted due to the varying language dataset sizes and difference in models. Statistical tests are hence skipped due to the cost involved in running parallel inference for all languages with varying LoRA modules.

\end{document}